\documentclass[runningheads]{llncs}
\usepackage[T1]{fontenc}
\usepackage{graphicx}
\usepackage{amssymb}
\usepackage{amsmath}
\usepackage{algorithm, algorithmic}
\usepackage{booktabs}
\usepackage{array}
\usepackage{adjustbox}
\usepackage{multirow}
\usepackage{xcolor}
\usepackage{verbatim}
\usepackage{marvosym} 
\usepackage{hyperref} 

\usepackage{hyperref}
\usepackage{color}

\begin{document}
\title{StepAL: Step-aware Active Learning for Cataract Surgical Videos}
\titlerunning{StepAL: Step-aware Active Learning for Cataract Surgical Videos}
%
\author{Nisarg A. Shah\inst{1}\inst{*}\Letter \and
Bardia Safaei\inst{1}\inst{*} \and
Shameema Sikder\inst{2,3} \and
S. Swaroop Vedula\inst{3} \and
Vishal M. Patel\inst{1}
}

%
\index{Shah, Nisarg A.}
\index{Sikder, Shameema}
\index{Vedula, S. Swaroop}
\index{Patel, Vishal M.}
\authorrunning{N. Shah et al.}
%
\institute{Johns Hopkins University, Baltimore, MD 21218, USA \and
Wilmer Eye Institute, Johns Hopkins University School of Medicine, Baltimore, MD \and 
Malone Center for Engineering in Healthcare, Johns Hopkins University \\ 
\email{snisarg812@gmail.com}}
\maketitle              
\def\thefootnote{*}\footnotetext{Equal contribution}
\begin{abstract}
Active learning (AL) can reduce annotation costs in surgical video analysis while maintaining model performance. However, traditional AL methods, developed for images or short video clips, are suboptimal for surgical step recognition due to inter-step dependencies within long, untrimmed surgical videos. These methods typically select individual frames or clips for labeling, which is ineffective for surgical videos where annotators require the context of the entire video for annotation. To address this, we propose StepAL, an active learning framework designed for full video selection in surgical step recognition. StepAL integrates a step-aware feature representation, which leverages pseudo-labels to capture the distribution of predicted steps within each video, with an entropy-weighted clustering strategy. This combination prioritizes videos that are both uncertain and exhibit diverse step compositions for annotation. Experiments on two cataract surgery datasets (Cataract-1k and Cataract-101) demonstrate that StepAL consistently outperforms existing active learning approaches, achieving higher accuracy in step recognition with fewer labeled videos. StepAL offers an effective approach for efficient surgical video analysis, reducing the annotation burden in developing computer-assisted surgical systems. 

\keywords{Cataract surgery \and Active Learning \and Step Recognition.}

\end{abstract}

\section{Introduction}

Automated surgical step recognition is critical for real-time surgical assistance \cite{padoy2019machine}, objective skill assessment \cite{yu2019assessment}, automated report generation \cite{zisimopoulos2018deepstep}, and improved training curricula \cite{funke2019video}. Annotated videos are necessary to develop algorithms for automated surgical step recognition, but reliable annotations are expensive because they require significant effort by trained experts \cite{shah2023glsformer,shah2025step,shah2025csmae,shah2025vision}. 

Techniques such as active learning (AL) \cite{safaei2025certainty,ren2021survey,Safaei_2025_WACV} can address the challenge of limited annotations by iteratively selecting the most informative and diverse surgical videos for annotation, minimizing labeling costs while maximizing model performance. Existing AL methods for recognition predominantly focus on image-level \cite{sener2017active,ma2024adaptive} or single-label short video clip classification \cite{sinha2019variational,taketsugu2024active}. Common strategies include uncertainty sampling \cite{wang2014new,wu2022entropy,safaei2024entropic}, diversity sampling \cite{sener2017active,ash2019deep,safaei2025filter}, heuristic approaches \cite{freund1997selective}, and ensemble models \cite{hino2023active}. 

While effective in their respective domains, existing AL techniques are not directly transferable to the complexities of long, multi-step surgical videos. A fundamental challenge is the granularity mismatch: frame- or clip-level selection conflicts with the practical need for complete, multi-step video annotation in surgical procedures. Due to the inherent sequential dependencies between surgical steps \cite{maier2017surgical,yu2019assessment,czempiel2020tecno}, individual clips often lack sufficient context for accurate labeling. As a result, partial video annotations are ineffective, as the entire video must be reviewed to ensure contextual accuracy. Standard AL methods typically operate directly on unlabeled training inputs, individual clips in the case of video recognition, which can lead to suboptimal performance for surgical videos that require step-level information embedded across sequential clips.

Standard AL strategies also often overlook the structural and temporal information implicitly available in pseudo-labels. While pseudo-labels may be imperfect, particularly in early AL cycles, they provide a valuable approximation of the step distribution within a video. This approximation offers a more informative selection signal than treating all clips equally, which is the implicit assumption in methods that rely solely on clip-level averaging. 
Furthermore, AL must account for both uncertainty and diversity, ensuring that the uncertainty-based selection process does not lead to redundant sample selection.

StepAL addresses these challenges with two key components. The Step-aware Feature Representation (SFR) captures inter-step dependencies by encoding the distribution of surgical steps within each video, leveraging pseudo-labels predicted by the step recognition model. This step-specific representation allows the selection process to effectively distinguish between different surgical videos based on their step composition. Complementing this, the Entropy-weighted Clustering (EWC) prioritizes videos exhibiting high overall uncertainty.  Critically, EWC leverages the step-aware representation of SFR, ensuring that the selected videos are not only uncertain but also represent a diverse range of surgical step sequences.  This combined approach ensures that StepAL focuses annotation efforts on videos that are both highly uncertain and representative of the diverse range of step sequences present in the dataset.

To the best of our knowledge, StepAL is the first AL framework specifically designed for video selection in the context of long, multi-step surgical video step recognition. It directly addresses the practical constraints and inherent sequential structure of surgical video data, surpassing the limitations of traditional frame- or clip-level AL approaches by jointly optimizing for uncertainty and diversity at the video level.

Our key contributions are:
\begin{itemize}
    \item We propose a novel active learning framework, StepAL, tailored for surgical video step recognition, effectively reducing the high annotation costs associated with labeling hour-long, untrimmed videos while achieving performance comparable to training using full annotations.
    \item Specifically, we introduce step-aware feature representations to effectively capture inter-step dependencies in surgical videos and an entropy-weighted clustering strategy to jointly prioritize videos with high model uncertainty and diverse surgical step distributions.
    \item Through extensive experiments on two publicly available cataract surgery datasets, we demonstrate the effectiveness of StepAL in enabling efficient and accurate surgical video step recognition.
\end{itemize}

\section{The StepAL Method}
\label{sec:stepal}

\begin{figure*}
    \centering
    \includegraphics[width=0.7\textwidth, trim={0.8cm 0.8cm 0.9cm 0.1cm}]{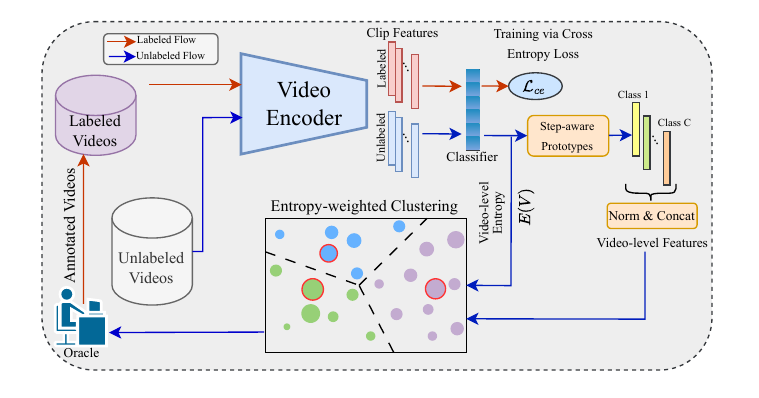}  
    \caption{Overview of our proposed \textit{StepAL} framework. Given long, untrimmed surgical videos as unlabeled data, StepAL employs a hybrid AL approach that selects informative samples based on both uncertainty and representativeness. Step-aware representations are obtained by concatenating prototypes from clip-level features of different pseudo-labels. Video-level uncertainty is measured by averaging clip-level entropies. Finally, entropy-weighted clustering selects videos closest to cluster centers, striking a balance between the diversity and uncertainty of the selected videos for annotation.} 
    \label{fig:framework} 
\end{figure*}

We introduce StepAL, an AL framework designed for efficient step recognition in long, untrimmed surgical videos. Our framework addresses the core challenge of minimizing annotation costs while maximizing model performance in multi-step procedures. 
Let \(\mathcal{D} = \{ V_n \}_{n=1}^{N}\) denote a surgical video dataset comprising \(N\) videos. We partition \(\mathcal{D}\) into a labeled set, \(\mathcal{D}_L\), and an unlabeled set, \(\mathcal{D}_U\). The AL process iteratively selects videos from \(\mathcal{D}_U\) for annotation by an expert (e.g., a surgeon) and adds them to \(\mathcal{D}_L\) to retrain the step recognition model.

\paragraph{Overall Pipeline.} Algorithm~\ref{alg:stepal} provides an overview of the StepAL procedure. The process begins by training a step recognition model, $F(\cdot;\theta)$, on the available labeled data, $\mathcal{D}_L$.  For each unlabeled video, $V \in \mathcal{D}_U$, we compute clip-level logits, $\ell_t \in \mathbb{R}^C$, and extract corresponding feature embeddings, $\phi_t \in \mathbb{R}^D$, where $C$ represents the number of distinct surgical steps and $D$ denotes the dimensionality of the feature space. Pseudo-labels, $\hat{y}_t = \arg\max_{c} (\ell_t)_c$, are generated for each clip based on the model's predictions.  These pseudo-labels are then used to construct a step-aware feature representation, $z_V \in \mathbb{R}^{C \times D}$ (detailed in Sec.~\ref{sec:sfr}), which captures the distribution of predicted steps within the video. Concurrently, a video-level entropy, $E(V)$, is computed by averaging the clip-level probability distributions and calculating the resulting entropy (Sec.~\ref{sec:ewc}).  This entropy serves as a measure of the model's overall uncertainty for the given video.  The core of the active learning selection strategy lies in applying weighted KMeans clustering to the set of step-aware feature representations, $\{z_V\}$, using the corresponding video entropies, $\{E(V)\}$, as sample weights.  This strategically biases the clustering towards videos exhibiting higher uncertainty.  A predefined budget, $b$, determines the number of videos selected; specifically, those closest to the cluster centers are chosen for full annotation. These newly annotated videos are then incorporated into the labeled set, \(\mathcal{D}_L\), and the unlabeled set, \(\mathcal{D}_U\), is updated accordingly. The entire process is repeated for a predetermined number of active learning cycles, $R$.

\begin{algorithm}[t]
\caption{StepAL: Active Learning for Multi-step Surgical Videos}
\label{alg:stepal}
\begin{algorithmic}[1]
\REQUIRE Dataset $\mathcal{D}$, initial labeled set $\mathcal{D}_L$, unlabeled set $\mathcal{D}_U$, total AL cycles $R$, budget per cycle $b$, number of classes $C$.
\FOR{cycle $r = 1$ \textbf{to} $R$}
    \STATE Train classifier $F(\cdot;\theta)$ on $\mathcal{D}_L$.
    \FOR{each video $V \in \mathcal{D}_U$}
        \STATE Infer clip-level logits $\ell_t \in \mathbb{R}^C$ and features $\phi_t \in \mathbb{R}^D$.
        \STATE Compute pseudo-labels $\hat{y}_t = \arg\max_{c}\, (\ell_t)_c$ for all clips $t$.
        \STATE Construct step-aware representation $z_V$ (Eq.~\ref{eq:step_sfr}).
        \STATE Compute video entropy $E(V)$ (Eq.~\ref{eq:video_entropy}).
    \ENDFOR
    \STATE Perform Weighted KMeans on $\{z_V\}$ with weights $\{E(V)\}$ (Eq.~\ref{eq:weighted_kmeans}).
    \STATE Select top-$b$ videos $\mathcal{Q}\subseteq \mathcal{D}_U$ nearest to each cluster center.
    \STATE Annotate all clips of videos in $\mathcal{Q}$; update $\mathcal{D}_L \leftarrow \mathcal{D}_L \cup \mathcal{Q}$ and $\mathcal{D}_U \leftarrow \mathcal{D}_U \setminus \mathcal{Q}$.
\ENDFOR
\RETURN Final labeled set $\mathcal{D}_L$ and trained model $F(\cdot;\theta)$.
\end{algorithmic}
\end{algorithm}

\subsection{Step-aware Feature Representation}
\label{sec:sfr}

Surgical procedures inherently consist of a sequence of distinct steps, each possessing unique visual and temporal characteristics. Traditional approaches that rely on simple feature averaging across all clips within a video discard this crucial step-specific information. To mitigate this limitation, we introduce a step-aware feature representation. This representation organizes clip-level embeddings according to their predicted surgical steps, thereby preserving the subtle, yet significant, differences between the various steps of a surgical procedure. This preservation of step-specific information is critical for enabling a more diverse and informative selection of videos during the active learning process.

For each unlabeled video, \(V \in \mathcal{D}_U\), composed of \(T\) clips \(\{x_1, x_2, \dots, x_T\}\), we extract clip-level features, \(\phi_t \in \mathbb{R}^D\), and generate corresponding pseudo-labels, \(\hat{y}_t \in \{1,\ldots,C\}\).  We define the set \(I_V^{(c)}\) as the indices of clips predicted to belong to step \(c\):

\begin{equation}
I_V^{(c)} \;=\; \bigl\{\,t \,\mid\, \hat{y}_t = c\bigr\}.
\end{equation}

For each surgical step \(c\), a step-specific feature, \(f_V^{(c)}\), is computed as follows:

\begin{equation}
f_V^{(c)} \;=\;
\begin{cases}
\displaystyle \frac{1}{|I_V^{(c)}|}\,\sum_{t \in I_V^{(c)}} \phi_t, & \text{if } I_V^{(c)} \neq \varnothing,\\
f_{\mathrm{average}}(V), & \text{otherwise},
\end{cases}
\end{equation}
where \(f_{\mathrm{average}}(V)\) represents the global average of \(\phi_t\) across all clips in video \(V\). This ensures that all surgical steps, even those not predicted in a particular video, are represented in the final feature vector. Each \(f_V^{(c)}\) is then \(\ell_2\)-normalized, and these normalized vectors are concatenated to form the final step-aware representation, \(z_V\):

\begin{equation}
\label{eq:step_sfr}
z_V \;=\;
\Bigl[\,
\tilde{f}_V^{(1)}
\;\|\;
\tilde{f}_V^{(2)}
\;\|\;
\dots
\;\|\;
\tilde{f}_V^{(C)}
\Bigr],
\quad
\tilde{f}_V^{(c)}
\;=\;
\frac{f_V^{(c)}}{\bigl\|\,f_V^{(c)}\,\bigr\|_2 + \epsilon}.
\end{equation}

The resulting step-aware representation, \( z_V \in \mathbb{R}^{C \times D} \), effectively encodes the distribution of predicted steps within each video.  By maintaining distinct feature representations for each predicted step, \(z_V\) captures the inherent compositional diversity of multi-step surgical procedures, a crucial factor for effective active learning.

\subsection{Entropy-weighted Clustering}
\label{sec:ewc}

In conjunction with the step-aware feature representation, we employ a strategy to prioritize videos with high model uncertainty. For each unlabeled video, \(V\), we quantify this uncertainty using a video-level entropy measure. Given the logit vector, \(\ell_t \in \mathbb{R}^C\), for clip \(t\), and its corresponding softmax probabilities, \(p_t = \mathrm{softmax}(\ell_t)\), we first compute the clip-level entropy:

\begin{equation}
H(p_t) = -\sum_{c=1}^C p_{t}^{(c)} \log(p_{t}^{(c)} + \epsilon).
\end{equation}

Then, the video-level entropy, \(E(V)\), is calculated by averaging the clip-level entropies across all \(T\) clips in the video:

\begin{equation}
\label{eq:video_entropy}
E(V) = \frac{1}{T} \sum_{t=1}^{T} H(p_t) =  -\frac{1}{T} \sum_{t=1}^{T} \sum_{c=1}^C p_{t}^{(c)} \log(p_{t}^{(c)} + \epsilon).
\end{equation}

High values of \(E(V)\) indicate that the model is uncertain about the step assignments within the video, often due to ambiguous surgical steps or complex transitions. Annotating such videos is expected to yield significant improvements in model performance.

To achieve a balance between uncertainty and diversity, we utilize weighted KMeans clustering.  This technique operates on the step-aware representations, \(\{z_V\}\), while incorporating the video entropies, \(\{E(V)\}\), as sample weights.  Let \(\{c_k\}_{k=1}^b \subset \mathbb{R}^{C\times D}\) denote the cluster centers, and \(\alpha(V)\) represent the cluster assignment for video \(V\).  The weighted KMeans objective function is:

\begin{equation}
\label{eq:weighted_kmeans}
\min_{\{c_k\}}
\;\sum_{V \,\in\, \mathcal{D}_U}\,
E(V)\,\bigl\|\,
z_V - c_{\alpha(V)}
\bigr\|^2.
\end{equation}
This formulation biases the clustering process towards videos with higher entropy values, effectively prioritizing the selection of uncertain samples.  Following the clustering process, we select up to \(b\) videos – those closest to each cluster center in the step-aware feature space – for full annotation. This selected set of videos represents a balance: high uncertainty (due to the entropy weighting) and diverse step compositions (due to the clustering on the step-aware representation). The annotated videos are then added to the labeled set, \(\mathcal{D}_L\), driving the iterative learning process.

The combination of step-aware feature representations and entropy-weighted clustering enables StepAL to efficiently identify videos that are both challenging for the current model and representative of the wide variety of surgical procedures.  This targeted approach to active learning minimizes the annotation effort while maximizing the information gain, ultimately leading to improved accuracy in surgical step recognition.

\section{Experiments and Results}
\noindent\textbf{Datasets:} We evaluate StepAL on two publicly available cataract surgery video datasets: 
Cataract-1k \cite{ghamsarian2024cataract} and Cataract-101 \cite{cataracts101}. The labeled subset of Cataract-1k provided by the authors includes 56 videos with a resolution of 1024x768 at 30 fps, annotated with 13 surgical steps. For this dataset, we use 25 videos for training, 7 for validation, and 24 for testing. Cataract-101 comprises 101 videos with a resolution of 720x540 at 25 fps, annotated with 10 surgical steps. We follow the standard split of 50 training, 10 validation, and 40 testing videos. Following prior work \cite{gao2021trans,twinanda2016endonet}, all videos are subsampled to 1 fps and resized to 250x250. Model performance is evaluated using frame-wise accuracy, precision, recall, and Jaccard Index. 

\noindent\textbf{Implementation Details:} We use a Video Vision Transformer (VideoViT) base model (`VideoViT-B/16` architecture \cite{arnab2021vivit}, pre-trained on Kinetics-400 \cite{kay2017kinetics}), with 16 frames sampled per video (16 x 224 x 224 input clips after resizing).  The model produces 768-dimensional feature embeddings ($\phi_t$) for each clip.

The active learning process starts with an initial labeled set, $\mathcal{D}_L$, containing 10\% of the training videos. The step recognition model, $F(\cdot; \theta)$, is fine-tuned on $\mathcal{D}_L$, then iteratively updated following Algorithm~\ref{alg:stepal}.  Each cycle ($R=4$ total) selects a new batch from the unlabeled set, $\mathcal{D}_U$, using the Step-aware Feature Representation (Sec. \ref{sec:sfr}) and Entropy-weighted Clustering (Sec. \ref{sec:ewc}), adding 10\% of the total training data to $\mathcal{D}_L$. This evaluates model performance up to 50\% labeled data utilization. A model trained on the \textit{complete} training set achieves mean accuracies of 92.01\% (Cataract-1k) and 89.47\% (Cataract-101), serving as an oracle performance reference.

Training uses a batch size of 14 on a single NVIDIA A100 GPU, employing the Adam optimizer \cite{kingma2014adam} (learning rate = 1e-5, $\beta_1 = 0.9$, $\beta_2 = 0.999$, weight decay = 5e-4) and Cross-Entropy Loss.

\noindent\textbf{Results:} Table \ref{tab:performance_metrics_combined} compares StepAL to state-of-the-art AL methods: Random, Margin \cite{balcan2007margin}, Entropy \cite{wang2014new}, Coreset \cite{sener2017active}, and CoreGCN \cite{caramalau2021sequential}.  Coreset focuses on diversity in feature space; CoreGCN uses a graph convolutional network.  StepAL, however, uniquely integrates both uncertainty and diversity via its step-aware feature representation (Sec. \ref{sec:sfr}) and entropy-weighted clustering (Sec. \ref{sec:ewc}).

At AL cycle $R=1$, StepAL outperforms all competing methods on both datasets. On Cataract-1k, StepAL achieves substantial improvements over the next best performing method, including a 4.66\% increase in accuracy and a 4.64\% increase in Jaccard index. Gains are also observed on Cataract-101, where StepAL surpasses the next best method by 1.23\% in accuracy and 1.00\% in Jaccard index, reflecting the dataset's inherent simplicity.

\begin{figure*}
    \centering
    \includegraphics[width=0.9\textwidth]{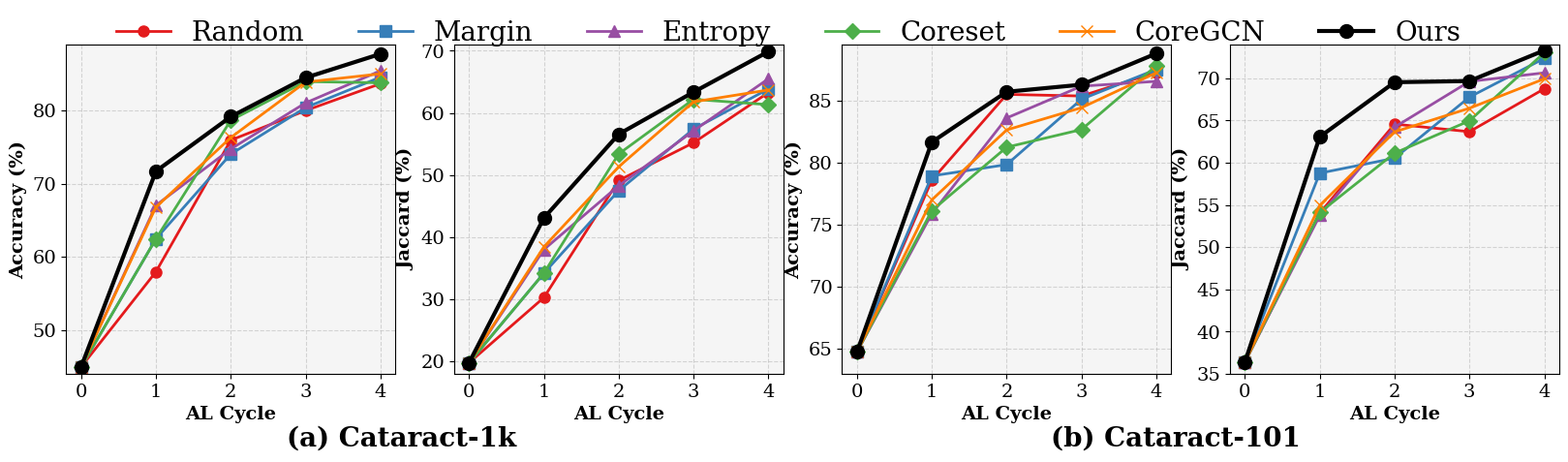}  
    \caption{Comparison of quantitative performance across 5 Active Learning Cycles (R = 0 to 4). (a) Results on Cataract-1k dataset and (b) Results on Cataract-101 dataset.}
    \label{fig:quant_performance}
\end{figure*}

Figure \ref{fig:quant_performance} shows StepAL's consistent advantage across all active learning cycles.  Its ability to achieve higher accuracy and Jaccard index from the initial stages highlights its effectiveness in rapidly identifying the most informative videos, crucial when annotation resources are limited.  On Cataract-101, performance converges at later cycles ($R=4$) as methods approach the oracle accuracy of 89.47\%; yet, StepAL maintains a performance edge throughout.

\begin{table}[ht]
\centering
\setlength{\tabcolsep}{4pt}
\renewcommand{\arraystretch}{0.9}
\resizebox{1.0\textwidth}{!}{
\begin{tabular}{llcccccc}
\toprule
\textbf{Dataset} & \textbf{Metric} & \textbf{Random} & \textbf{Margin}\cite{balcan2007margin} & \textbf{Entropy}\cite{wang2014new} & \textbf{Coreset}\cite{sener2017active} & \textbf{CoreGCN}\cite{caramalau2021sequential} & \textbf{Ours} \\
\midrule
\multirow{4}{*}{Cataract-1k} 
  & Accuracy  & 0.5795 & 0.6245 & \textit{0.6703} & 0.6245 & 0.6679 & \textbf{0.7169 \textcolor{teal}{(+4.66\%)}} \\
  & Precision & 0.5074 & 0.5299 & 0.5706       & 0.5299 & \textit{0.5868} & \textbf{0.6485 \textcolor{teal}{(+6.17\%)}} \\
  & Recall    & 0.4691 & 0.5008 & \textit{0.5277} & 0.5008 & 0.5242 & \textbf{0.5785 \textcolor{teal}{(+5.08\%)}} \\
  & Jaccard   & 0.3028 & 0.3420 & 0.3801       & 0.3420 & \textit{0.3844} & \textbf{0.4308 \textcolor{teal}{(+4.64\%)}} \\
\midrule
\multirow{4}{*}{Cataract-101}
  & Accuracy  & 0.7859 & \textit{0.7893} & 0.7589 & 0.7613 & 0.7700 & \textbf{0.8016 \textcolor{teal}{(+1.23\%)}} \\
  & Precision & 0.6937 & \textit{0.7495} & 0.7002 & 0.7100 & 0.7300 & \textbf{0.7635 \textcolor{teal}{(+1.40\%)}} \\
  & Recall    & 0.6791 & \textit{0.7314} & 0.6891 & 0.7040 & 0.7054 & \textbf{0.7333 \textcolor{teal}{(+0.19\%)}} \\
  & Jaccard   & 0.5404 & \textit{0.5877} & 0.5376 & 0.5411 & 0.5495 & \textbf{0.5977 \textcolor{teal}{(+1.00\%)}} \\
\bottomrule
\end{tabular}
}
\caption{Performance Metrics for Two Cataract Surgery Datasets, Cataract-1k~\cite{ghamsarian2024cataract} and the Cataract-101 dataset \cite{cataracts101} for R = 1. Values in the green indicate the absolute percentage increase (from the next best result).}
\label{tab:performance_metrics_combined}
\end{table}

\noindent\textbf{Ablations:} Our ablation study on Cataract-1k, summarized in Table \ref{tab:ablation_metrics}, illustrates the effectiveness of StepAL's components. The \textit{Random} baseline serves as a benchmark, with the \textit{Entropy} method improving accuracy by 15.7\% over Random by selecting videos based on average clip-level entropy, demonstrating the value of incorporating uncertainty into active learning. In contrast, KMeans underperforms Entropy by 7.3\% due to its reliance on averaged clip features, which obscure essential details. However, \textit{ME-KMeans} (Maximum Entropy KMeans), which also uses averaged features but selects the most uncertain video in each cluster, surpasses both Entropy and KMeans, showing the importance of combining diversity with uncertainty for effective selection..

EWC shows only marginal improvement over KMeans and still lags behind ME-KMeans, emphasizing that the feature representation is a critical component. In stark contrast, \textbf{Ours} (StepAL) integrates step-aware feature representation (Sec. \ref{sec:sfr}) and EWC (Sec. \ref{sec:ewc}), outperforming all other methods by significant margins. StepAL not only improves accuracy by 5.3\% over ME-KMeans but also enhances precision, demonstrating its robustness through consistent performance improvements across all metrics, effectively capturing step-level diversity and prioritizing overall video uncertainty for more effective video selection.

\begin{table}[ht]
\centering
\setlength{\tabcolsep}{4pt}
\renewcommand{\arraystretch}{0.9}
\resizebox{0.75\textwidth}{!}{
\begin{tabular}{lcccccc}
\toprule
\textbf{Metric} & \textbf{Random} & \textbf{Entropy} & \textbf{KMeans} & \textbf{ME-KMeans} & \textbf{EWC} & \textbf{Ours} \\
\midrule
Accuracy  & 0.5795 & 0.6703 & 0.6245 & {0.6807} & 0.6408 & \textbf{0.7169} \\
Precision & 0.5074 & 0.5706 & 0.5299 & {0.6157} & 0.5366 & \textbf{0.6485} \\
Recall    & 0.4691 & 0.5277 & 0.5008 & {0.5317} & 0.5123 & \textbf{0.5785} \\
Jaccard   & 0.3028 & 0.3801 & 0.3420 & {0.3941} & 0.3491 & \textbf{0.4308} \\
\bottomrule
\end{tabular}
}
\caption{Performance Metrics for Ablation using the Cataract-1k\cite{ghamsarian2024cataract} dataset.}
\label{tab:ablation_metrics}
\end{table}
\section{Conclusion}

In this paper, we present StepAL, a novel AL framework designed for the selection of videos to improve surgical step recognition.
Unlike traditional AL methods that focus on individual frames or clips, StepAL selects entire videos for annotation, aligning better with real-world surgical workflows.
Our approach combines step-aware feature representation, which captures fine-grained step-level information using pseudo-labels, with entropy-weighted clustering. This method prioritizes videos that are both highly uncertain and diverse in their step composition for further labeling. Experiments on two cataract surgery datasets demonstrate that StepAL consistently outperforms existing active learning approaches, achieving higher accuracy in step recognition with fewer labeled videos.

\subsection*{Acknowledgements.}
This work was supported by a grant from the National Institutes of Health, USA; R01EY033065. The content is solely the responsibility of the authors and does not necessarily represent the official views of the National Institutes of Health.

\bibliographystyle{splncs04}
\bibliography{Paper-3578}

\end{document}